# VAE-LIME: Deep Generative Model Based Approach for Local Data-Driven Model Interpretability Applied to the Ironmaking Industry

Cedric Schockaert[†]
Paul Wurth S.A.
Luxembourg, Luxembourg
cedric.schockaert@paulwurth.com

Vadim Macher
Ecole Supérieur d'Ingénieurs
Léonard de Vinci
Courbevoie, France
vadim.macher@edu.devinci.fr

Alexander Schmitz
Paul Wurth S.A.
Luxembourg, Luxembourg
alexander.schmitz@paulwurth.com

## ABSTRACT

Machine learning or deep learning applied to generate black-box data driven-models are lacking of transparency leading the process engineer to lose confidence in relying on the model predictions to optimize his industrial process. Bringing processes in the industry to a certain level of autonomy using data-driven models is particularly challenging as the first user of those models, is the expert in the process with often decades of experience. Therefore, it is necessary to expose to the process engineer, not solely the model predictions, but also the interpretability for each prediction. Several approaches have been proposed in the literature to make trained models interpretable. The Local Interpretable Model-agnostic Explanations (LIME) method has gained a lot of interest from the research community recently. The principle of this method is to train a linear model that is locally approximating the black-box model, by generating randomly artificial data points locally. Model-agnostic local interpretability solutions based on LIME have recently emerged to improve the original method. We present in this paper a novel approach, VAE-LIME, for local interpretability of data-driven models forecasting the temperature of the hot metal produced by a blast furnace. Such ironmaking process data is characterized by multivariate time series with high inter-correlation representing the underlying process in a blast furnace. Our contribution is to use a Variational Autoencoder (VAE) to learn the complex blast furnace process characteristics from the data. Consequently, the VAE is aiming at generating optimal artificial samples to train a local interpretable model better representing the black-box model in the neighborhood of the input sample processed by the black-box model to make a prediction. In comparison with LIME, VAE-LIME is showing a significantly improved local fidelity of the local interpretable linear model with the black-box model resulting in robust model interpretability.

## 1 Introduction and background

Humanity is moving towards a data-driven world where data is the decisional core for any industrial process. The large amount of generated data is a trigger for establishing complex data-driven black-box models that act at different level of an industrial organization in order to provide a certain level of autonomy for process control. In the recent years, machine learning and particularly deep learning models have been applied successfully to solve various problems, and therefore tend to support or even replace human in various decisional tasks. However, the predictive accuracy reached by deep learning models, as a consequence of their significantly higher number of parameters, has as drawback a lake of interpretability, leading the process expert to a subjective choice to trust or not the generated predictions. Indeed, the acceptance level to put such a model in production is based solely on the error statistics evaluated during the validation phase, without any justifications for each prediction the model is providing as results. Lacking a justification for the black-box data-driven model prediction, is leading the domain expert to be unable to understand or extrapolate the model behavior for any possible operation of his process. In the process industry, domain experts having years of experience are often reluctant in the acceptance of a black-box data-driven model because of a lack of its interpretability.

By definition, the interpretability of a data-driven black-box model is the ability of the model to provide any insight about the output it is generating, allowing the domain expert to trust the model. Interpretability is also a requirement for model validation before its deployment in production, and for the validation of its output when deployed in production, where the interpretability is providing extra dimensions from which the domain expert can derive rules for accepting or not the underneath model predictions.

Model interpretability or explainable AI (XAI) is a research field that is gaining significantly increasing interest since few years [1, 2]. Several approaches have been proposed in order to discover the hidden justification of any output generated by a black-box model. Three groups of model interpretability approaches can be derived from the state-of-the-art:

**Example-based**: the interpretability of a data-driven black-box model output for a specific input is provided by listing similar inputs that have been used for training the model. For the process industry, this could be for example the recognition of a specific process operation close to one used to train the model.

**Global**: such approaches are providing global interpretability of a model, and don't explain each output generated by the model. Those methods are however very interesting to classify a model in the process industry according to the target process operation for which the model has been potentially optimized by selecting specific data for training. Global model interpretability is acting as a signature generation of black-box models.

**Local**: those approaches are providing an explanation for each output generated by the model, and therefore are allowing an instance based model interpretability [3, 4, 5].

Several methods for model interpretability have been proposed in the literature with different applications. It is important to make a distinction between model-agnostic and model specific approaches. A model-agnostic approach, on the contrary to a model-specific approach, is a method that is independent to the algorithm used to train the data-driven black-box model to interpret, and therefore acts as a generic procedure to open any black-box model. The predominant advantage is that it is a post-hoc method, therefore any existing trained model can be made interpretable, and there are no constraints in the selection of the algorithm to train the data-driven black-box model. Indeed, there exists intrinsic explainable algorithm like Cart [6] or linear regression, but they are lacking of predictive power due to their inherent low complexity providing biased predictions. Two prominent methods are covered currently in research: saliency or perturbation based. Saliency based methods [7], are aiming to build salience map for neural networks by input gradient calculation [8]. Perturbation based methods are quite intuitive [9]. For example, in a popular method called LIME [9], it is assumed that a linear interpretable model acting as a surrogate model, can locally approximate the data-driven black-box model. For that purpose, perturbations are generated around the input utilized by the data-driven black-box model to generate a prediction requiring to be interpreted. LIME has gained a lot of popularity since 2016 and represents today a reference algorithm for model-agnostic local interpretability. Another popular approach is based on the calculation of the Shapley values [10], however that method has as drawback the long processing time due to the underlying simulation of coalitional game theory.

The research community has recently proposed improvements of LIME [11, 12]. In [11], a hierarchical clustering approach is first applied to create clusters that will drive the perturbation generation to reduce the inherent interpretability variation of LIME induced by the randomness of perturbations. The approach proposed in [12] is improving the stability of the interpretability by using an autoencoder to select most relevant perturbations randomly generated. Both approaches are based on a regularized random perturbations selection.

In this paper, we are proposing a new approach for local model interpretability based on LIME, where the generation of perturbations is performed by a generative deep learning model, a Variational AutoEncoder (VAE) [13]. VAEs have been implemented for various applications in fake image generation [14], but also new discovery in multiple fields [15]. The VAE is aiming at generating significantly more representative perturbations of the underlying process for training the local interpretable surrogate model. This is providing a better stability of the interpretability while improving the local fidelity of the local surrogate model with respect to the data-driven black-box model to interpret. In the following section, results are presented for the interpretability of a data-driven model predicting the temperature of the hot metal produced by a blast furnace [16]. Those results are benchmarked with the traditional LIME approach. Conclusion and perspectives of this research are discussed.

## Description of the proposed approach

Autoencoders (AE) [17] are trained to encode an input in a latent space with lower dimension. The decoder is aiming at reconstructing that input from its compressed representation in the latent space. During the training phase, the Mean Square Error (MSE) between the input and its reconstruction is minimized. AEs are acting as features extractor as only the relevant input characteristics are preserved in the latent space. By definition, AEs are not suited for content generation as there is no regularization of the latent space during the training phase. Indeed, a regularized latent space exhibits properties, like spatial continuity, allowing a meaningful reconstruction of any random point located in that space. The distance between points in the latent space is related to their similarity. By definition, AEs are trained to reach overfitting in order to ensure a minimum reconstruction loss. A VAE is an AE trained with a specific regularization term in the loss function to ensure that the latent space has the required properties for an optimal generative process. To enable this regularization, VAEs have a modified encoding-decoding process where an input is encoded as a normal distribution over the latent space, and not as a single point. The training procedure of a VAE is schematized in Figure 1, where the regularization term of the loss function is the Kullback-Leiber divergence (KL) penalizing the encoding of the input in a distribution that is not following a standard normal distribution. As a consequence, a spatial correlation in the latent space is reached after convergence during the training phase.

VAE-LIME, as presented in Figure 1, is using as sample generator, a VAE trained on the same training dataset as the black-box model to be interpreted for a test input $x_{test}$. N random samples are generated in the latent space of the VAE. Those samples are generated from a gaussian distribution where the mean $x^l_{test}$ is the representation of $x_{test}$ in the latent space, and $\sigma_j$ is the standard deviation for each dimension *j* of the latent space. The number N of samples and $\sigma_j$ for each dimension *j* of the latent space are the parameters of VAE-LIME. For each generated sample *i* in [1,…,N], a weight $w_i$ is calculated as being the complement of the Gower distance [18] between that sample position in the latent space, and the mean $x^l_{test}$. Each sample is reconstructed by the decoder of the VAE and an output per sample is generated by the black-box model. Finally, a weighted linear regression model is applied to the set of samples $s_i$ and associated outputs $y_i$ using weights $w_i$, in order to provide the local variable importance for the black-box model output corresponding to the test input $x_{test}$. The variable importance is the associated coefficient of the linear regression.

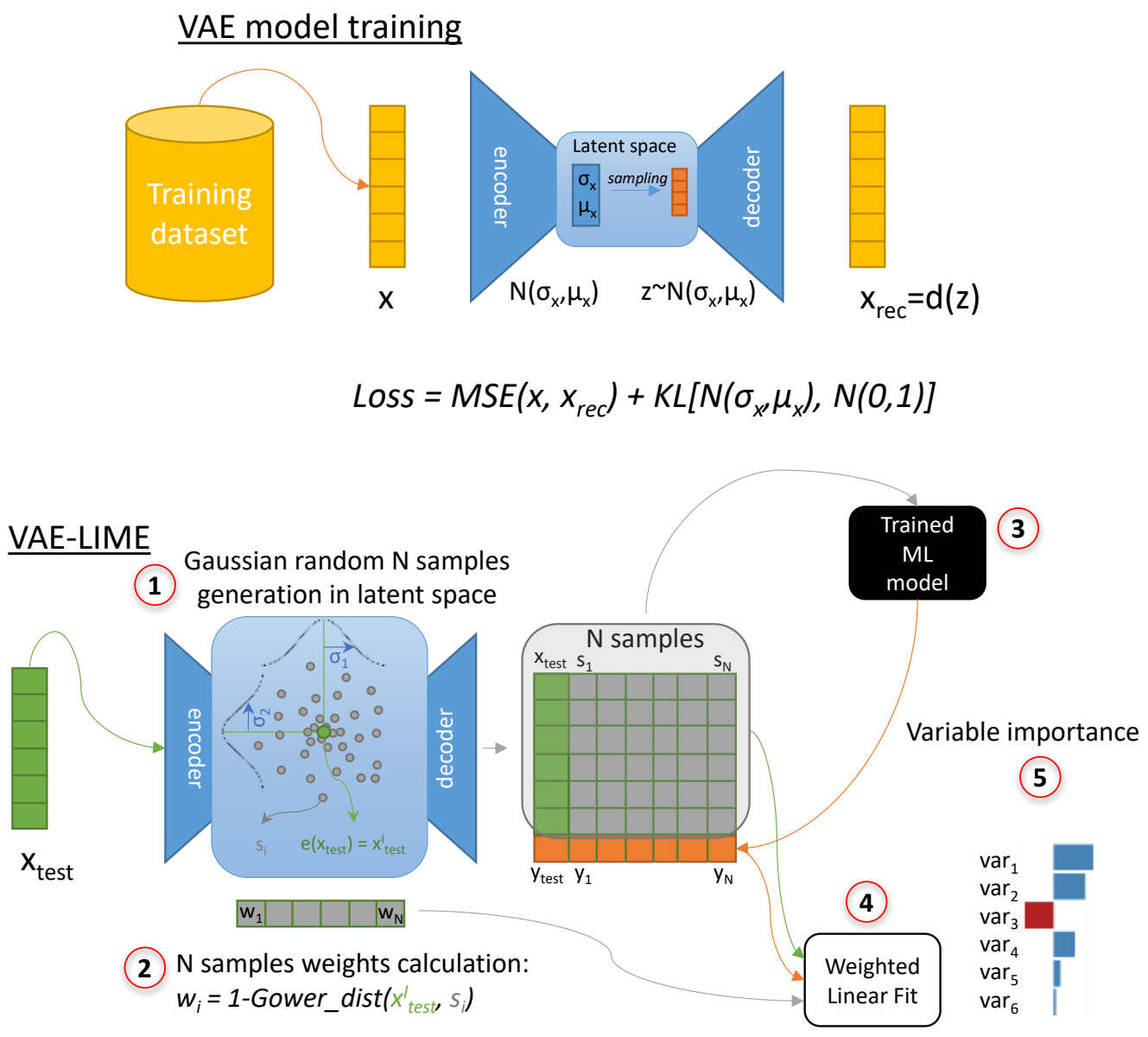

**Figure 1: Architecture of VAE-LIME. Top: the training procedure of a VAE using the training dataset of the black-box model to locally interpret; bottom: VAE-LIME algorithm**

On the contrary to other approaches proposed as an improvement to LIME, VAE-LIME is controlling the generation of samples, and not filtering randomly generated samples. This characteristic of VAE-LIME is aiming at providing a better local interpretability of the black-box model by providing a prediction with the local linear model close to the prediction of the black-box model for the test sample $x_{test}$.

Figure 2 illustrates results for one black-box model predicting the temperature of the hot metal produced by a blast furnace. The ten most important variables given by VAE-LIME and LIME are compared. Figure 2c,d are illustrating, for both methods, the scatter plot between the weights calculated for each generated samples and the corresponding predictions from the black-box model. This is providing a visual assessment of the stable generation of samples in VAE-LIME having weight values more uniform compared to LIME. Figure 2e summarizes the statistics of the model interpretability for the evaluated test sample. The statistics are further discussed with the illustration of Figure 3.

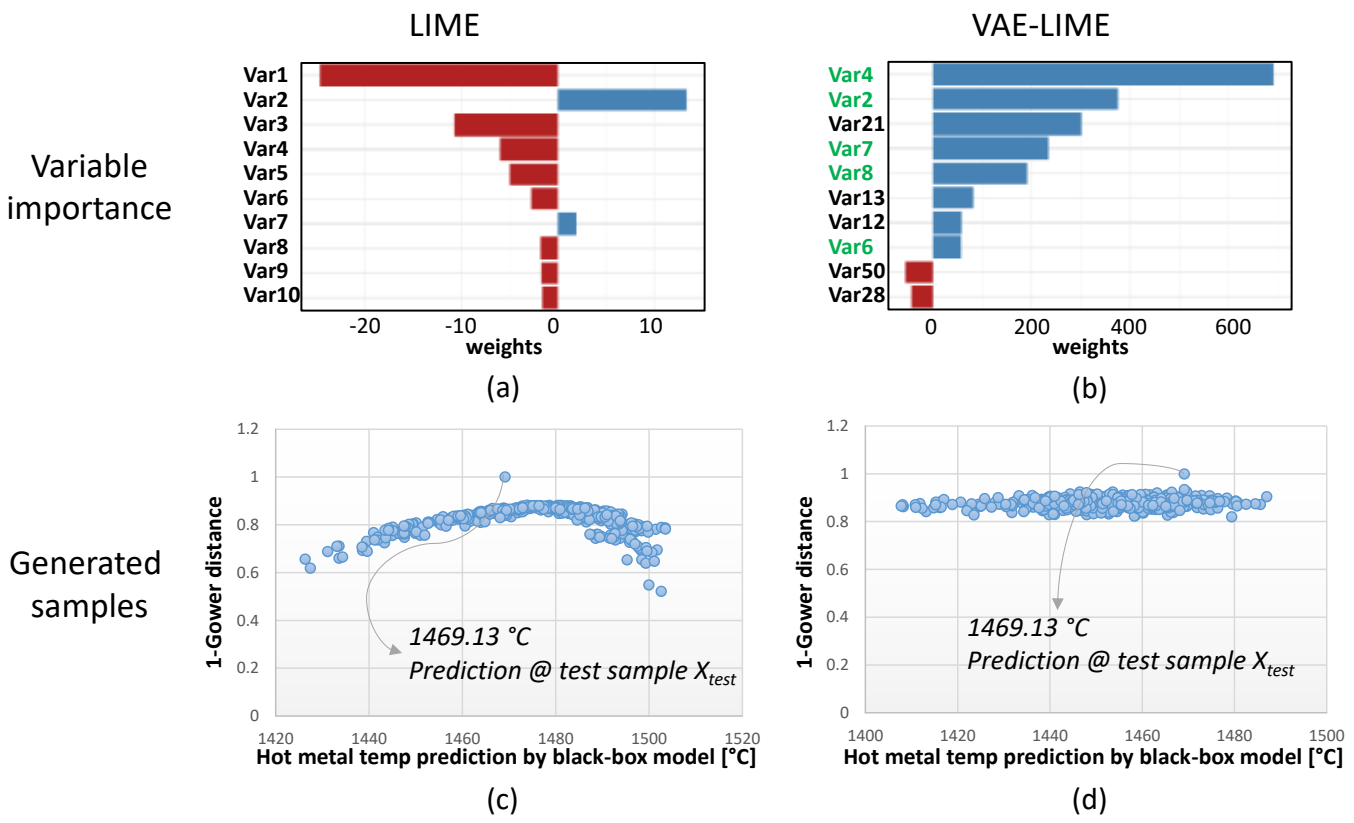

Weighted Linear Fit

| metric | LIME | VAE-LIME |
|---|---|---|
| R2 (explanation fit) | 0.93 | 0.97 |
| MSE | 19.4 | 6.1 |
| Abs err @$X_{test}$ | 0.57 | 0.005 |

(e)

**Figure 2: VAE-LIME vs LIME comparison of the variable importance for one test sample. Variables commonly selected by both approaches are highlighted in green.**

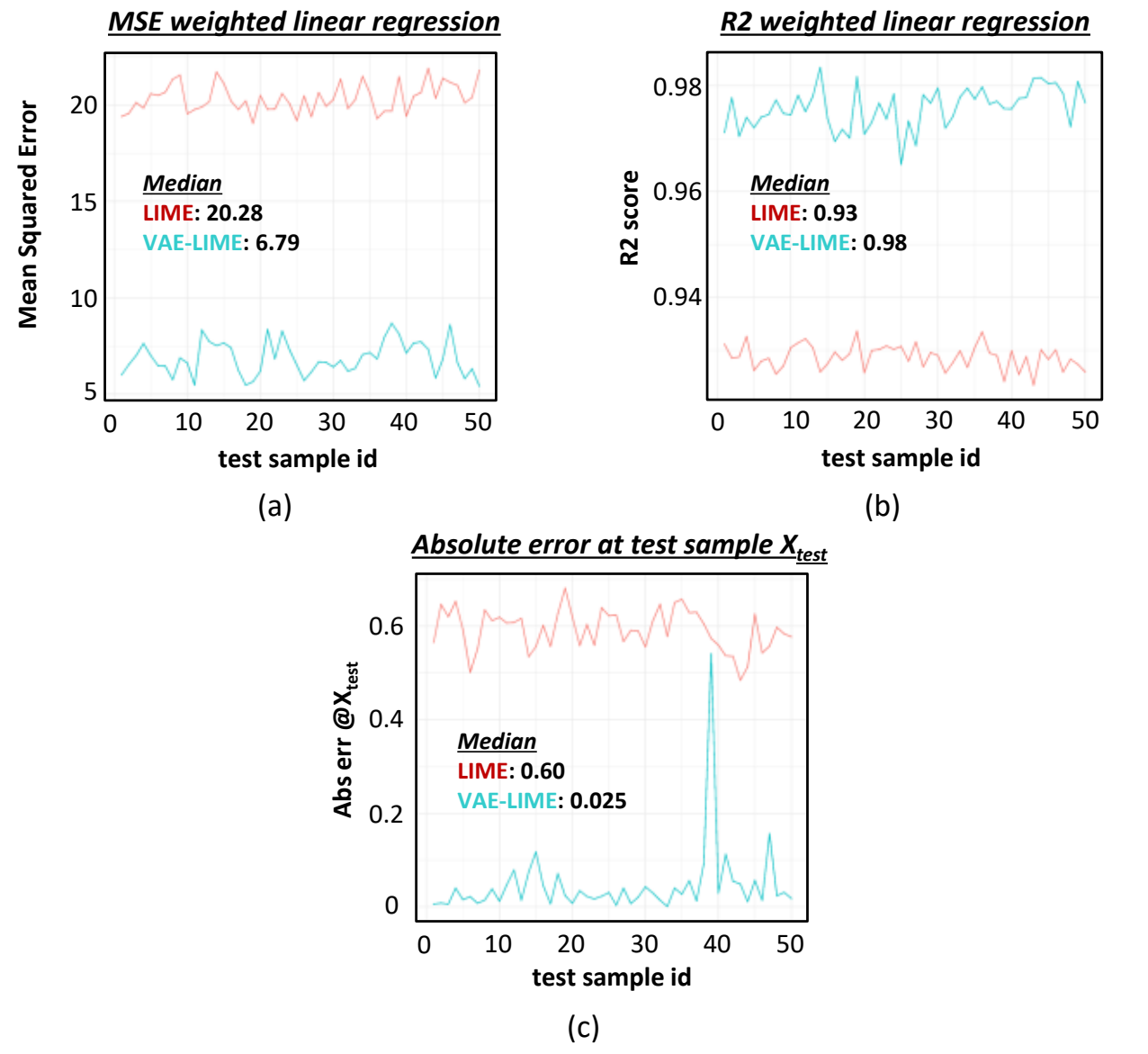

**Figure 3: VAE-LIME results validation benchmarked with LIME for a test set of 50 predictions to interpret. (a) MSE evolution; (b) R2 score evolution; (c) absolute error evolution between the linear and black-box model predictions**

Those results are validated in Figure 3 and compared to LIME for 50 test samples. The validation focuses first on the fidelity of the local linear model, with respect to the black-box model around each test sample. For that purpose, the Mean Square Error (MSE) between the predictions of the local linear model and the black-box model predictions is calculated (Figure 3a) for each test sample including their corresponding N generated samples. The fidelity of the local linear model is significantly improved compared to LIME. This is further confirmed by analyzing in Figure 3c the absolute error between the predictions given by the local linear model and the black-box model at the test sample input $x_{test}$. The fidelity improvement of the local linear model with the black-box model is the objective of VAE-LIME where the sample generation is restricted by the learned inter-variable correlation induced by the underlying blast furnace process. For LIME, on the other hand, the N samples are generated randomly regardless to any relations between the variables.

The R2 score of the local linear model is providing a measurement of the confidence in the variable importance for the local linear model. Only when the fidelity level, as measured with the MSE, of the linear model with the black-box model is high, then R2 becomes a measurement of the confidence for the black-box model interpretability. As illustrated in figure 3b, the R2 score of VAE-LIME is improved compared to LIME.

## 3 Conclusion and perspectives

The research conducted in this paper is aiming at improving the local fidelity of LIME with respect to the black-box data-driven model to interpret. For that purpose, a Variational Autoencoder is implemented to generate the data with higher fidelity with respect to the underlying process in the blast furnace. A major issue with LIME for the interpretability of our model predicting the temperature of the hot metal produced by a blast furnace, is the randomization of the generation of samples for the local interpretability.

By controlling the generation of samples by using a deep generative network, the local MSE between the linear interpretable model and the black-box model has been significantly improved compared to LIME. As a consequence of this, the absolute error between the prediction of the linear and the prediction to interpret provided by the black box model for a specific input, is reduced outstandingly.

An extension of this research is aiming at further evaluate the proposed approach by developing complementary metrics to reflect the time stability of the variable importance. Indeed, the blast furnace process is characterized by a high inertia, leading to a certain stability depending on the current operation. The predictive model for the hot metal temperature has captured this inertia but with some potential limitations in relation with the data used for training and other considerations to be taken into account to characterize prediction bias of data-driven models. Therefore, the black-box model to interpret carries a certain temporal smoothness for the variable importance of consecutive predictions.

The validation of the local interpretability of a black-box data-driven model by a post-hoc method based on a local surrogate model must be carefully handled, and this is a major reason why opening the ‘black box’ is an ongoing research subject and few products in the market are offering this option to customers. The continuity of our research is covering the construction of a benchmark to assess complementary solutions. As a result, comparison of variable importance of several algorithms will provide insights about their relative behavior. However, an absolute reference is a requirement for properly conclude a first research phase and propose this as product to process engineers. Another step we will initiate to go in that direction is to apply VAE-LIME to interpret predictions made by an interpretable model by nature, for example a tree based model. The danger is to make early optimistic conclusions, the model to interpret having by definition a lower complexity. Another path to explore is to train very specific models dedicated to one particular operation of the blast furnace, and therefore being by construction highly sensitive to a short list of variables. Those models have therefore a signature that can be used to validate the local interpretability to some extent.